%% file: main.tex
\pdfoutput=1
\documentclass[11pt,a4paper]{article}
\usepackage[hyperref]{emnlp2020}
\usepackage{times}
\usepackage{latexsym}
\usepackage{tikz}
\usepackage{pgfplots}
\usepackage{amssymb}
\usepackage{multirow}
\usepackage{booktabs}
\usepackage{enumitem}
\usepackage{physics}

\usepackage{comment}
\usepackage{microtype}

\definecolor{yelloworange}{RGB}{179, 143, 0}

\definecolor{lightpurple}{RGB}{185, 139, 232}
\definecolor{lightpink}{RGB}{252, 113, 196}
\definecolor{darkcyan}{RGB}{66, 215, 244}
\definecolor{myorange}{rgb}{1,0.5,0}
\definecolor{darkgreen}{RGB}{110, 160, 88}
\definecolor{myred}{RGB}{160, 70, 49}
\definecolor{mypink}{RGB}{253, 170, 255}
\definecolor{mypurple}{RGB}{120, 33, 122}
\definecolor{darkyellow}{RGB}{176, 188, 83}
\definecolor{darkgrey}{RGB}{160, 160, 155}
\definecolor{brightgreen}{RGB}{131, 191, 47}

\newenvironment{tightitemize}{\itemize\addtolength{\itemsep}{-8pt}}{\enditemize}

\aclfinalcopy 
\def\aclpaperid{2596} 

\pgfplotsset{compat=1.16}

\title{Compositional Demographic Word Embeddings}

\author{Charles Welch, Jonathan K. Kummerfeld, Verónica Pérez-Rosas, \and Rada Mihalcea  \\
Computer Science \& Engineering \\
University of Michigan \\
\texttt{\{cfwelch,jkummerf,vrncapr,mihalcea\}@umich.edu}
}

\date{}

\begin{document}
\maketitle
\begin{abstract}
Word embeddings are usually derived from corpora containing text from many individuals, thus leading to general purpose representations rather than individually personalized representations. While personalized embeddings can be useful to improve language model performance and other language processing tasks, they can only be computed for people with a large amount of longitudinal data, which is not the case for new users. We propose a new form of personalized word embeddings that use demographic-specific word representations derived compositionally from full or partial demographic information for a user (i.e., gender, age, location, religion). We show that the resulting demographic-aware word representations outperform generic word representations on two tasks for English: language modeling and word associations. We further explore the trade-off between the number of available attributes and their relative effectiveness and discuss the ethical implications of using them.
\end{abstract}

\section{Introduction}

Word embeddings are used in many natural language processing tasks as a way of representing language.
Embeddings can be efficiently trained on large corpora using methods like word2vec or GloVe~\cite{mikolov2013efficient,pennington-etal-2014-glove}, which learn one vector per word.
These embeddings capture syntactic and semantic properties of the language of all individuals who contributed to the corpus.
However, they are unable to account for user-specific word preferences (e.g., using the same word in different ways across different contexts), particularly for individuals whose usage deviates from the majority. These individual preferences are reflected in the word's nearest neighbors. As an example, Table~\ref{tab:examples_change} shows the way two users use the word ``health'' and the word's five nearest neighbors in their respective personalized embedding spaces. The word is used in similar contexts, where contextual embeddings may give similar representations, but it has different salient meanings in the personal space of each user. User A tends to talk more about preventative care and insurance, while user B tends to talk about people's experiences affecting their mental health.

The typical approach in natural language processing (NLP) is to use one-size-fits-all language representations, which do not account for variation between people.
This may not matter for people whose language style is well represented in the data, but could lead to worse support for others \cite{pavalanathan-eisenstein-2015-confounds,may-etal-2019-measuring,kurita2019measuring}.
While the way we produce language is not a direct consequence of our demographics or any other grouping, it is possible that by tailoring word embeddings to a group we can more effectively model and support the way they use language.

Additionally, personalized embeddings can be useful for applications such as predictive typing systems that auto-complete sentences by providing suggestions to users, or dialog systems that follow the style of certain individuals or professionals (e.g., counselors, advisors). They can also be used to match the communication style of a user, which would signal cooperation from a dialog agent. 

\begin{table}
    \small
    \centering
    \setlength{\tabcolsep}{5pt}
    \begin{tabular}{p{5mm} p{3.0cm} p{3.3cm}}
        \toprule
        User &
        Example Use &
        Nearest Neighbors \\
        \midrule
        A
        & doctors think this is bad for her \textbf{health} ...
        & preventative, insurance, reform, medical, education
        \\[4pt]
        B
        & it is usually bad for your \textbf{health} ...
        & professional, mental, conduct, experiences, online
        \\[4pt]
        All & N/A &
        medical, preventative, insurance, safety, healthcare
        \\
        \bottomrule
    \end{tabular}
    \caption{\label{tab:examples_change}
    Nearest neighbors of the word ``health'' for two different users in personalized and a generic embedding space.
    }
\end{table}

In this paper, we propose compositional demographic word embeddings as a way of building personalized word embeddings by leveraging data from users sharing the same demographic attributes (e.g., age: {\it young}, location: {\it Europe}). Our proposed method has the benefits of personalized word representations, while at the same time being applicable to users with limited or no data. 

To implement and evaluate our proposed method, we build a large corpus of Reddit posts from 61,981 users for whom we extract self-reported values of up to four demographic properties: age, location, gender, and religion.  
We examine differences in word usage and association captured by the demographics we extracted and discuss the limitations and ethical considerations of using or drawing conclusions from this method.
We explore the value of compositional demographic word embeddings on two English NLP tasks: language modeling and word associations. In both cases, we show that our proposed embeddings improve performance over generic word representations.

\section{Related Work}
\label{sec:related-work}

\paragraph{Embedding Bias.}
Recent work on embeddings has revealed and attempted to remove racial, gender, religious, and other biases~\cite{manzini-etal-2019-black,bolukbasi2016man}.
The bias in our corpora and embeddings have a societal impact and risks exclusion and demographic misrepresentation~\cite{hovy-spruit-2016-social}. This means that users of certain regions, ages, or genders may find NLP technologies more difficult to use. For instance, when using standard corpora for POS tagging, \citet{hovy-sogaard-2015-tagging} found that models perform significantly lower on younger people and ethnic minorities. Similarly, results on text-based geotagging show best results for men over 40~\cite{pavalanathan-eisenstein-2015-confounds}.

Similar results are starting to be found in embeddings produced by contextual embedding methods \cite{may-etal-2019-measuring,kurita2019measuring}. We focus on non-contextual embedding methods because of their computational efficiency, which is crucial if many separate representations are being learned. Additionally, there may not be a large amount of available data for underrepresented groups and these contextualized models require billions of tokens for training. Recent work has also shown that static embeddings are competitive with contextualized ones in some settings~\cite{arora2020contextual}.

\paragraph{Personalization.}
The closest work is \citet{garimella-etal-2017-demographic}'s exploration of demographic-specific word embedding spaces.
They trained word embeddings for male and female speakers who live in the USA and India using skip-gram architectures that learn a separate word matrix for each demographic group (e.g., male speakers from the USA).

Another line of work used discrete~\cite{hovy-2015-demographic} or continuous values~\cite{lynn-etal-2017-human} to learn speaker embeddings: a single vector for each user.
The speaker embedding is appended to the input of the recurrent or output layer, and trained simultaneously with the rest of the model.
This idea applies to any contextual information type and was introduced as a way to condition language models on topics learned by topic modeling \cite{mikolov2012context}.
It has since been used as a way of representing users in tasks such as
task-oriented and open-domain dialog \cite{wen2013recurrent,li-etal-2016-persona},
information retrieval based on book preferences \cite{amer2016toward},
query auto-completion \cite{jaech-ostendorf-2018-personalized},
authorship attribution \cite{ebrahimi2016personalized},
sarcasm detection \cite{kolchinski-potts-2018-representing},
sentiment analysis \cite{zeng2017socialized},
and cold-start language modeling \citet{huang2016enriching}.
Finally, a recent study by \citet{king-cook-2020-evaluating} compared how to improve a language model with user-specific data using priming and interpolation, depending on the amount of data available, learning a new model for each user.

More generally, personalization has been extensively applied to marketing, webpage layout, product and news recommendation, query completion, and dialog~\cite{eirinaki2003web,das2007google}.
\citet{Welch19LearningFromPersonal,Welch19LookWhosTalking} explored predicting response time, common messages, and speaker relationships from personal conversation data.
\citet{zhang2018personalizing} conditioned dialog systems on artificially constructed personas and \citet{madotto-etal-2019-personalizing} used meta-learning to improve this process.
Goal-oriented dialog has used demographics (i.e. age, gender) to condition system response generation, showing that this relatively coarse grained personalization improves system performance \cite{joshi2017personalization}.

\paragraph{Social Media.}
We use social media data with demographic attributes inferred from user posts.
Prior work has explored extraction or prediction of attributes such as age, gender, region, and political orientation~\cite{rao2010classifying,rangel2013overview}.
Work on analyzing the demographics of social media users also includes race/ethnicity, income level, urbanity, emotional stability, personality traits \cite{mccrae1997personality}, and life satisfaction \cite{duggan2013demographics,correa2010interacts}.

One particularly relevant study by \citet{gjurkovic-snajder-2018-reddit} presented a corpus of Reddit users with personality information as well as some demographics for a subset of users.
Unlike our approach, which is based on text content, they extract information from Reddit flairs, a type of user tag.
Out of their set of 10,295 users, 2,253 are also in our set of users (22\% of theirs, 0.5\% of ours) that have one or more demographic labels, confirming the speculation in their paper that extracting demographics from text is a complementary approach that captures more information about users in their data.
Other work has used Reddit posts to identify users who were diagnosed with depression \cite{yates-etal-2017-depression} and to construct personas for personalized dialog agents \cite{mazare-etal-2018-training}.

\paragraph{Language Models.}
To evaluate embeddings, we consider language modeling, a task that has long been used for speech recognition and translation, and more recently been widely used for model pretraining.
A range of models have been developed, with progressively larger models trained on more data~\cite[e.g.,][]{dai-etal-2019-transformer}.
Variations of the LSTM have consistently achieved state-of-the-art performance without massive compute resources
\cite{merity2018analysis,melis2019mogrifier,merity2019single,li-etal-2020-learning}.
We use the AWD-LSTM~\cite{merity2018regularizing} in our experiments, as it achieves very strong performance, has a well-studied codebase, and can be trained on a single GPU in a day.

\section{Dataset}\label{sec:dataset}

Our first contribution is a new dataset. We use English Reddit comments as they are publicly available, are written by many users, and span multiple years.\footnote{\url{https://www.reddit.com/r/datasets/comments/3bxlg7/i\_have\_every\_publicly\_available\_reddit\_comment/}}
We extract demographic properties of users from self-identification in their text.

\subsection{Finding Demographic Information}
\label{sec:demographics}

Reddit users do not have profiles with personal information fields that we could scrape.
Instead, we developed methods to extract demographic information from the content of user posts.

In order to determine what kind of information we can extract about users, we performed a preliminary analysis.
We manually labeled a random sample of 132 statements that users made about themselves. We specifically searched for statements starting with phrases such as `i am a' or `i am an'. In our sample:
36\% clearly stated the user's age, religion, gender, occupation, or location;
34\% contained descriptive phrases that were difficult to categorize like `i am a big guy' or `i am a lazy person'; and
30\% mentioned attributes such as sexual orientation, dietary restrictions, political affiliations, or hobbies that were rare overall.

Based on our analysis, we decided to focus on age, religion, gender, occupation, and location as the main attributes.\footnote{We attempted to extract occupations, but found they were difficult to identify and group because there are many different occupations, many ways of stating one's occupation, and many ways to describe the same occupation.} These were extracted as follows:

\noindent
\textbf{Age}. We extracted the user's age using a regular expression.\footnote{.*?(i am\textbar i\textbackslash 'm) (\textbackslash\textbackslash d+) (years\textbar yrs\textbar yr) old[\textasciicircum e].*?} During this process, we found users that were matched to different ages due to the corpus covering user activity across several years. In those cases, we removed users whose age difference was greater than the time span of our corpus. Additionally, we excluded users who said they were less than 13 years of age, as this violates the Reddit terms of service.
We decided to split the age into two groups, young and old at a threshold of 30, as this split was used in previous work~\cite{rao2010classifying}, and it gave a reasonable split for our data and the data we used for testing word associations~\cite{garimella-etal-2017-demographic}.

\noindent
\textbf{Gender.} Gender was extracted by searching for statements referring to oneself as a `boy', `man', `male', `guy', for male, or `girl', `woman', `female', `gal', for female. 
Manual inspection revealed some users indicated that they were of both genders. In that case, if one gender occurred less than one fifth of the time we took the majority of the reported gender, otherwise we removed the user from our dataset.
We acknowledge that this approach excludes transgender, gender fluid, and a range of non-binary people, and may misgender people as well (see \S~\ref{sec:ethics} for further discussion of these issues).

\noindent
\textbf{Location.} To obtain location information, we searched for phrases such as `i am from' and `i live in.' Next, whenever either the next token is (1) tagged as a location by a named entity recognizer \cite{manning-EtAl:2014:P14-5}, (2) a noun, or (3) the word `the', we select all subsequent tokens in the phrase as the user location.
Manual inspection of matches showed that Reddit users are not consistent in the granularity of reported location. Statements included cities, state, province, country, continent, or geographical region.
Based on the number of users per country, we decided to merge some countries into region labels while leaving others separate.
This resulted in the following set of regions: USA, Asia, Oceania, UK, Europe, Canada. We further matched location statements to lexicons to resolve the location to one of these regions, removing common relative location words.\footnote{northern, western, eastern, southern, downtown, suburbs} For larger population regions of Canada and the USA, we match statements using state abbreviations, province names, highest population cities, and in the USA we also match the capital cities. For other regions we only match the highest population cities as there were too many cases to cover.

\noindent
\textbf{Religion.} To extract religion, we searched for the five largest global religious populations,\footnote{From \url{https://www.adherents.com/}, although note that since our study the domain name has been hijacked by a payday loans service. The site is archived by the Library of Congress at \url{https://www.loc.gov/item/lcwaN0003960/}} counting `secular', `atheist', and `agnostic' as one non-religious group. We used a regular expression\footnote{.*?(i am\textbar i\textbackslash 'm) (a )?(christian \textbar\  muslim \textbar\  secular \textbar\  atheist \textbar\  agnostic \textbar\  hindu \textbar\  buddhist).*?} and filtered users who stated beliefs in more than one of these five groups.

\subsection{Post-processing}

The resulting dataset was further filtered to remove known bots.\footnote{\url{https://www.reddit.com/r/autowikibot/wiki/redditbots}} For the demographic data we consider two subsets.
First, the set of users for which all four attributes are known (4Dem).
With this set we perform ablation experiments on the number of known attributes in a controlled manner.
However, it is important to note that this set may not be representative of most users on Reddit, as it focuses on users willing to divulge a range of demographic attributes.
Our second sample addresses this by including users for whom we identify two or more of the demographic attributes (2+Dem).
Statistics for these sets are described in Table \ref{tab:corpus_stats}, along with the training, development, and test splits used for the language modeling experiments. 

\begin{table}
    \centering
    \small
    \begin{tabular}{lrr}
        \toprule
        Set & Users & Posts \\
        \midrule
        All Reddit & 13,213,172 & 1,430,935,783 \\
        \midrule
        \multicolumn{3}{c}{2+Dem}\\
        \midrule
        Total & 61,627 & 205,394,970 \\
        Training & 34,110 & 50,000 \\
        Validation & 9,190 & 10,000 \\
        Test & 9,143 & 10,000 \\
        \midrule
        \multicolumn{3}{c}{4Dem} \\
        \midrule
        Total & 354 & 3,433,062 \\
        Test & 354 & 10,000 \\
        \bottomrule
    \end{tabular}
    \caption{
    Statistics for two Reddit sets: with at least two demographic attributes (2+Dem), or all four demographic attributes (4Dem). Training, development, test splits used in the language modeling experiments are also shown. First row shows overall number of posts and users from the entire set of Reddit posts.}
    \label{tab:corpus_stats}
\end{table}

\input{demographic_distributions.tex}

\input{different_words_no_nn.tex}

The distribution of demographic values for each of these sets is shown in Figure \ref{fig:distribution_attributes}.
Looking at the set of all users in our data who have at least two known demographic attributes (2+Dem), we find that 83\% of the time location is unknown. Age and religion are the next most frequently missing at 53\% and 34\% respectively.
Gender is more likely to be known than the other attributes: only 10\% of users in this subset have an unknown gender.
In a manual evaluation of all our extracted attribute labels for the 100 users, we found accuracies of 94\% for location and gender, 78\% for religion, and 96\% for age. Additional details of this evaluation are provided in the supplementary material.

\section{Generating Compositional Demographic Word Embeddings}\label{sec:generating_embeddings}

We propose two methods for learning compositional demographic embeddings.
The first learns a generic embedding for each word and a vector representation of each demographic attribute (including `unknown').
This is memory efficient, as we need only 19 vectors to cover all of our attributes.
In the second method, for each word we learn (a) a generic embedding and (b) a vector for each demographic attribute.
This is more expressive, but requires twenty vectors for each word.

\subsection{Demographic Attribute Vectors}\label{sec:vectors}

In this approach we jointly learn a matrix for words and a separate vector for each demographic value.
The word matrix $W \in \mathbb{R}^{|V|\times k}$ has a row for each word in the vocabulary and a k-dimensional vector for each embedding. The demographic values can be represented by another matrix $D \in \mathbb{R}^{|C|\times k}$, where $C$ is the set of all demographic values (e.g., male, female, christian, USA). The hidden layer is calculated as $h=W_h^\intercal(W_w + C_g + C_l + C_r + C_a)$ where $w$ represents the one-hot encoding of an input word and $g,l,r,a$ represent the demographic values of the speaker. This is a modified skip-gram architecture~\cite{mikolov2013efficient} with a hierarchical softmax, which sums five terms so that back-propagation updates the word representation as well as the demographic values.

We use posts from all users to train embeddings for words that occur at least five times across all users. This yields a vocabulary of 503k words. We learn 100-dimensional embeddings with an initial learning rate of 0.025 and a window size of five.

\subsection{Demographic Word Matrices}\label{sec:matrices}

When learning demographic matrices we separately run our skip-gram model for each of the demographic attributes (e.g., gender) and learn a generic word matrix $W_G \in \mathbb{R}^{|V|\times k}$ and a value specific word matrix for each value, $v$, of the given attribute, $A$, (e.g., male, female) $W_v \in \mathbb{R}^{|V|\times k},  \forall v \in A_v$. This changes the hidden layer calculation to $h=W_h^\intercal {G}_{w} + W_h^\intercal {W_v}_{w}$, with hidden layer weights $W_h$, and the model then learns a generic word representation, in matrix $G$, while learning the value specific impact on the meaning of that word.

\paragraph{Differences Across Demographic Embeddings.}
In order to understand what our embeddings capture, we examine words that have different representations across demographics. We can look at the nearest neighbors of a given query word across the embedding spaces for different demographics. We perform this analysis on both the demographic matrices and vectors, finding less variation in the neighbors when using demographic vectors, making them less interesting.
We show examples of words with low overlap in nearest neighbors for demographic matrices in Table \ref{tab:word_differences}. These show the differences in word meaning across groups.

\section{Language Modeling}

We first examine the usefulness of our embeddings by showing that they can help us better model a user's language.
We consider two experiments.
First, we focus on compositional demographic embeddings and sample 50k posts from our corpus for training the language model and 5k for each of validation and test.
Next, we compare with a user-specific model on a sample of our data with text from just 100 users who each have a large amount of data available in our corpus, with an average of 3.2 million tokens per user.

In both experiments, we use the language model developed by \citet{merity2018regularizing,merity2018analysis}.
As discussed in \S~\ref{sec:related-work}, this model was recently state-of-the-art and has been the basis of many variations.
We modify it to initialize the word embeddings with the ones we provide and to concatenate multiple embedding vectors as input to the recurrent layers.
The rest of the architecture is unaltered. We tried adding rather than concatenating and found no improvement. We chose to concatenate the inputs with the intuition that the network would learn how to combine the information itself.

We explored various hyperparameter configurations on our validation set and found the best results using dropout with the same mask for generic and demographic-specific embeddings, untied weights, and fixed input embeddings.
Untying and fixing input embeddings is supported by concurrent work \cite{emnlp20lm}.
Each model is trained for 50 epochs.
We use the version from the epoch that had the best validation set perplexity, a standard metric in language modeling that measures the accuracy of the predicted probability distribution.

\subsection{Demographic Perplexity Evaluation}

Table~\ref{tab:ppl_on_50krand} shows results for our demographic personalization methods, which are designed to handle new users for whom we have demographics but not much text data.
The first method, demographic vectors, performs no better than generic embeddings.
This is surprising since prior work has achieved success on a range of tasks with this kind of representation (see \S~\ref{sec:related-work}).
We suspect that for language modeling the variations are too fine-grained to be captured by a single vector. However, demographic matrices do improve significantly over generic embeddings.
A model with all demographics improves the most, but we also see improvements when only one demographic value is known.

The LSTM hidden layer size is the same across models, but the change in the input size affects the total number of parameters. To control for this, we ran our baseline model and model initialized with generic words with a larger input size, matching the number of parameters in our best models. As shown in Table~\ref{tab:ppl_on_50krand}, this increase in parameters does not improve performance.

\begin{table}
    \centering
    \small
    \begin{tabular}{lcc}
        \toprule
        Model Type and Input Size & 2+Dem & 4Dem \\
        \midrule
        Baseline, 100 & 123.8 & 124.6 \\
        Baseline, 500 & 125.1 & 126.1 \\
        Generic Words Only, 100 & 116.0 & 112.1 \\
        Generic Words Only, 500 & 115.8 & 112.6 \\
        Demographic Vectors, 200 & 116.7 & 113.0 \\
        Demographic Matrices & & \\
         + Age Only, 200 & 109.4 & 110.3 \\
         + Gender Only, 200 & 109.4 & 109.9 \\
         + Location Only, 200 & 109.7 & 112.9 \\
         + Religion Only, 200 & 110.9 & 112.0 \\
         + All Demographics, 500 & {\bf 107.7} & {\bf 109.1} \\
        \bottomrule
    \end{tabular}
    \caption{\label{tab:ppl_on_50krand}
    Perplexity on the demographic data.
    Our demographic-based approach improves performance.
    The difference between the last row and generic words is significant ($p<0.00001$ with a permutation test).
    }
\end{table}

\input{ppl_breakdown.tex}

\subsubsection{Ablation Experiments} 
Table~\ref{tab:ppl_on_50krand} shows results when using no demographics (top 4 rows), one demographic at a time (rows 6-9) and all four demographics (row 10).
Each attribute improves perplexity, with age and gender improving it more than location and religion.

Additionally, we perform a breakdown of the performance of our demographic matrices language model on each of the demographic groups. These results are shown in Table \ref{tab:ppl_demo_breakdown}. We do see worse performance on some minorities as compared to other groups for the same model, although that is not always the case (gender, for instance, shows better perplexities for female than for male, and Muslim shows lower perplexity than Christianity, which has substantially more data). When we use the demographic word embeddings in our model, we are able to improve performance for all demographic groups, including minorities.

We also find that the performance on the `unknown' group increases in all cases with our largest improvement on `unknown' religion. The unknown is explicitly modeling people in our dataset who have either (1) stated this demographic information with a value that we model but not in a way that our regular expressions identify, (2) stated this demographic information with a value that we do not model, or (3) have not stated this demographic information. In the second case, the effect is that it is useful to know which demographic groups the speaker does not belong to. In the third case, it may be that not sharing this particular piece of information (while sharing other personal information) says more about what the speaker will tend to say.

\subsection{Comparison with User Representations}

For users with a lot of data, it is possible to train a user-specific model, with embeddings that capture their unique language use.
We would expect this to be better than our demographic embeddings, but also only be feasible for users with a lot of data.
This experiment compares our demographic approach with a user-specific approach.

We create a model for each user using the sample that has a large amount of data for 100 users (3.2 million tokens each on average) as done in concurrent work~\cite{coling20personal}.
We tried two approaches, user vectors and user matrices, which are analogous to our demographic vectors and matrices.
The difference is that rather than having a separate vector / matrix for each demographic we have a separate vector / matrix for each user. Our split sizes for language model experiments are the same as the demographic experiments.

\paragraph{Results.}
Table~\ref{tab:results_all} shows results for generic embeddings, user vectors, user matrices, and demographic matrices.
We find that user vectors, as have been used widely in previous work~\cite{kolchinski-potts-2018-representing,li-etal-2016-persona}, do not improve performance.
Both our demographic and user matrices improve performance over generic embeddings with comparable performance.
While we chose 100 users with a lot of data, they had less data than the amount used to train each demographic specific model.
The relationship between the amount of data, its similarity to a user's writing, and the effect on performance is an interesting open question.

\begin{table}
    \centering
    \small
    \begin{tabular}{l c}
        \toprule
        Model                         & PPL \\
        \midrule
        Generic Word Embeddings   & 63.94 \\
        User Vectors              & 68.98 \\
        User Matrices             & 61.69 \\
        Demographic Matrices      & 61.80 \\
        \bottomrule
    \end{tabular}
    \caption{\label{tab:results_all}
    Comparing our demographic-based approach with two user-specific approaches.
    Perplexities are generally lower than previous tables because the threshold for rare words being made UNK was higher.
    }
\end{table}

\section{Demographic Word Associations}

\input{demo_aware_associations.tex}

\input{demo_aware_8subsets.tex}

As a second evaluation, we consider word associations, a core task in NLP that probes the relatedness or similarity between words.
Data is collected for the task by presenting a stimulus word (e.g., \emph{cat}) and asking people what other words come to mind (e.g., \emph{dog} or \emph{mouse}).
Earlier systems relied on resources such as WordNet to solve the task, but most recent work has used word embeddings.

\paragraph{Data.}
For our evaluation, we use data from \citet{garimella-etal-2017-demographic}.
They constructed a word association dataset and experimented with learning separate word embedding matrices for different demographic groups.
To collect the data, they (1) asked crowd workers to write one word associated with a single word prompt and (2) asked the workers their gender, age, location, occupation, ethnicity, education, and income.
Only gender and location information was released, but the authors provided age information upon request.

\paragraph{Evaluation.}
As in prior work, we consider evaluation metrics defined in terms of:
$f_w$, the number of people who listed word $w$ for a stimulus;
$f_{max}$, the highest $f_w$ across all words chosen for a stimulus; and
$t$, the number of participants given a stimulus.

\noindent \emph{best} is $f_w$ divided by $f_{max}$, where $w$ is the word in the embedding space closest to the stimulus word;
\emph{ooN} (out-of-N) is $\sum f_w/t$ for the \emph{N} words in the embedding space closest to the stimulus word; 
both are averaged over all stimulus words.

We consider two experiments.
One directly matches \citet{garimella-etal-2017-demographic}, testing each demographic group separately.
Since our interest is in compositionality, we also introduce a setting where the data is split into eight disjoint sets, one for each combination of the three attributes.

\paragraph{Models.}
\citet{garimella-etal-2017-demographic} proposed two methods, which we merge by taking the best result from either one.
We considered only our demographic matrix embeddings as they performed best on language modeling.
For the experiment with separate demographics, we use the appropriate embeddings.
For the experiment with combinations of demographics, we concatenate the embeddings.
We also compare to concatenation of generic embeddings learned for each attribute (this performs better than any individual generic embedding).

\paragraph{Results.}
Table~\ref{tab:demo_associations} shows results on the single-demographic experiment.
We achieve higher performance, but that may come from the change in training dataset.\footnote{
Their models are trained on 67.6m tokens of blog data, while ours are trained on 1,400m tokens of Reddit data.
}$^,$\footnote{
We see a larger gain for the US than the IN evaluation. This may be because in our data location is unknown for many users and India is underrepresented (so much so that we aggregate it into all of Asia).
}
Table~\ref{tab:demo_association_combinations} shows results on the multi-demographic setting.
We include only the best pair (age and gender) due to space.
We have seen in earlier experiments that location does not perform as well as the other attributes and found the same trend here.
Overall, composing demographic-based representations helps, with a combination of all three attributes consistently performing well on the \emph{oo3} metric, while having two helps on the \emph{best} metric.
Generic embeddings only score the highest on one subset: Male, India, Young.

\section{Limitations and Ethical Considerations} \label{sec:ethics}

This work uses demographic information to modify language representation.
This type of work is encouraged by the numerous arguments outlined in \cite{perez2019invisible}, which demonstrate the need for demographic data disaggregation in order to make decisions and build technologies that are equitable for all. 
We view our work as an initial investigation of differences in language model performance across demographics and how technology can be improved for the identified groups.
Our results in Tables~\ref{tab:ppl_on_50krand} and~\ref{tab:ppl_demo_breakdown} show that using demographic information can enable the development of language tools that improve performance for all groups compared to simply training on all data.

Although we show that some language production aspects are correlated with demographic information, we do not believe the way we speak is a direct and only consequence of one's demographics, neither do we claim that this is the ideal information source for it or that this will necessarily hold for populations sampled significantly differently than in our study. As a consequence, it is possible that using demographics in embedding construction could accentuate bias, although this remains to be studied. Those that use our method should account for this possibility.

Our study uses four demographic variables and only covers a subset of the potential values of each demographic. For instance, we do not use the same granularity across locations, include all locations, religions, or gender identities. We simplify age into ranges. The groups `secular', `agnostic', and `atheist' are grouped into one broader group. Our sample is further biased by the choice of platform as each platform contains text from different populations. Users in our sample are predominately young, male, atheist, and live in the United States.

When using gender as a study variable, we followed the recommendations of \newcite{larson-2017-gender}. Our ``gender" extraction method does not refer to biological sex. 
After running gender extraction patterns, users are assigned to either the `male', `female', or `unknown' label, meaning that on the basis of these phrases one's gender identity is assumed to be binary or to be a gender identity unknown to our model, which may include those who are transgender, non-binary, or those who do not wish to disclose their gender. However, we are aware that the use of regular expressions for the extraction of demographic attributes can lead to false positives and false negatives (error rates are provided in the supplemental material) and that there exists a bias in using these strategies, as populations that do not wish to be identified are less likely to explicitly make such statements. For transparency, our released code includes the scripts used to assign demographic labels.

Above we discussed concerns for incorrect demographic assignment when developing models.
There are also potential negative consequences when using these models in a deployed system.
Our embeddings can only be used when the demographics of a user are known.
This may be acceptable if the user voluntarily self-reports their demographics with the understanding that they will alter the predictions they receive.
However, if demographics are automatically inferred there is a risk of misattribution, which depending on the application may have negative consequences.

A separate consideration is the environmental impact of this approach.
Compared to the standard method, our approach does involve training more models, but the cost of inference is likely only marginally higher. We believe the additional cost in training is worth the benefits to individual users.

Finally, we acknowledge that components of our method could potentially be used for user profiling~\cite{rangel2013overview} and/or surveillance of target populations, thus exposing members of underrepresented groups to harms such as discrimination and coercion and threatening intellectual freedom~\cite{richards2012dangers}. Similarly, the language models could be used to generate text in the style of a target population or at least to estimate the label distribution of a given text, which would help obfuscate the identity of the author~\cite{potthast2018overview}. This obfuscation could help hide an author's identity in order to avoid surveillance or could be used maliciously to infiltrate communities online. 
We advocate against the use of our methods for these or other ethically questionable applications.

\section{Conclusions}

We proposed a novel method of generating word representations by composing demographic-specific word vectors.
Through experiments on two core language processing tasks, language modeling and word associations, we show that demographic-aware word representations outperform generic embeddings. 
We also find that demographic matrices perform much better than demographic vectors.
Through several ablation analyses we show that word embeddings that leverage multiple demographic attributes give better performance than those using single attributes.

To support future work that can help model individuals and demographics, our code is available at \url{http://lit.eecs.umich.edu}. Our data is not available due to licensing restrictions but can be redownloaded and processed with our scripts. We hope this will support work on solutions for NLP applications and resources that can better serve minorities and underrepresented groups.

\section*{Acknowledgements}

We would like to thank Saif Mohammad, Heng Ji, Mona Diab, Allison Lahnala, Thamar Solorio, and the anonymous reviewers for their helpful suggestions.
This material is based in part on work supported by IBM (Sapphire Project), DARPA (grant \#D19AP00079), Bloomberg (Data Science Research Grant), the NSF (grant \#1815291), and the John Templeton Foundation (grant \#61156). Any opinions, findings, conclusions, or recommendations in this material are those of the authors and do not necessarily reflect the views of IBM, DARPA, Bloomberg, the NSF, or the John Templeton Foundation. 

\bibliography{emnlp2020}
\bibliographystyle{acl_natbib}

\appendix

\input{supplemental.tex}

\end{document}

%% file: demographic_distributions.tex
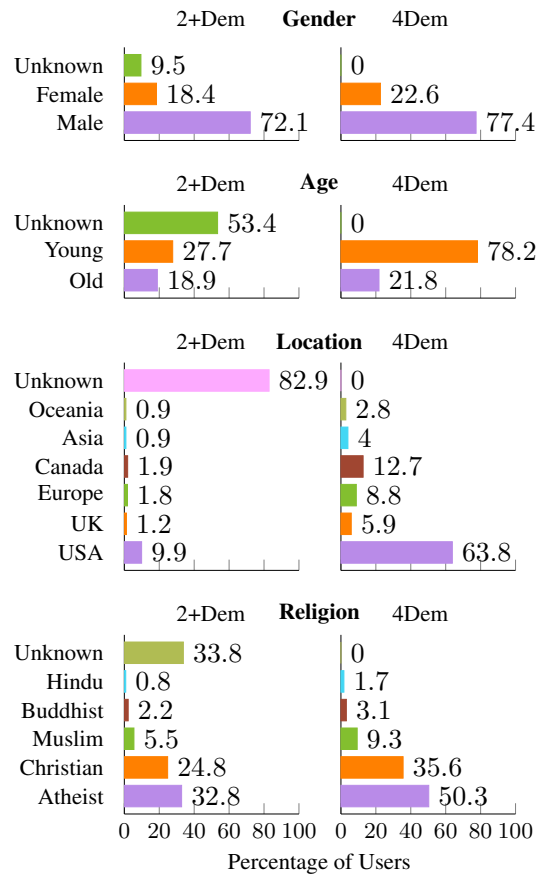
\begin{figure}
    \centering
    \scalebox{0.95}{
    \begin{tikzpicture}
        \begin{axis}[
            width=.25\textwidth,
            xbar=0pt,
            /pgf/bar shift=0pt,
            legend style={
            legend columns=4,
                at={(xticklabel cs:0.5)},
                anchor=north,
                draw=none
            },
            ytick={0,...,5},
            ytick style={draw=none},
            axis y line*=none,
            axis x line*=bottom,
            tick label style={font=\footnotesize},
            legend style={font=\footnotesize},
            label style={font=\footnotesize},
            bar width=3mm,
            yticklabels={
                {Male},
                {Female},
                {Unknown},
            },
            xticklabels={},
            xmin=0,
            xmax=100,
            area legend,
            y=4mm,
            enlarge y limits={abs=0.625},
            nodes near coords,
            nodes near coords style={text=black},
            every axis plot/.append style={fill}
        ]
        \addplot[lightpurple] coordinates {(72.1,0)};
        \addplot[myorange] coordinates {(18.4,1)};
        \addplot[brightgreen] coordinates {(9.5,2)}; 
        \end{axis}
        
        \begin{axis}[
            width=.25\textwidth,
            xshift=3.0cm,
            xbar=0pt,
            /pgf/bar shift=0pt,
            legend style={
            legend columns=4,
                at={(xticklabel cs:0.5)},
                anchor=north,
                draw=none
            },
            ytick={0,...,5},
            ytick style={draw=none},
            axis y line*=none,
            axis x line*=bottom,
            tick label style={font=\footnotesize},
            legend style={font=\footnotesize},
            label style={font=\footnotesize},
            xtick={0,20,...,100},
            bar width=3mm,
            yticklabels={},
            xticklabels={},
            xmin=0,
            xmax=100,
            area legend,
            y=4mm,
            enlarge y limits={abs=0.625},
            nodes near coords,
            nodes near coords style={text=black},
            every axis plot/.append style={fill}
        ]
        \addplot[lightpurple] coordinates {(77.4,0)};
        \addplot[myorange] coordinates {(22.6,1)};
        \addplot[brightgreen] coordinates {(0.0,2)};
        \end{axis}
        
        \begin{axis}[
            width=.25\textwidth,
            yshift=-2.2cm,
            xbar=0pt,
            /pgf/bar shift=0pt,
            legend style={
            legend columns=4,
                at={(xticklabel cs:0.5)},
                anchor=north,
                draw=none
            },
            ytick={0,...,5},
            ytick style={draw=none},
            axis y line*=none,
            axis x line*=bottom,
            tick label style={font=\footnotesize},
            legend style={font=\footnotesize},
            label style={font=\footnotesize},
            xtick={0,20,...,100},
            bar width=3mm,
            yticklabels={
                {Old},
                {Young},
                {Unknown},
            },
            xticklabels={},
            xmin=0,
            xmax=100,
            area legend,
            y=4mm,
            enlarge y limits={abs=0.625},
            nodes near coords,
            nodes near coords style={text=black},
            every axis plot/.append style={fill}
        ]
        \addplot[lightpurple] coordinates {(18.9,0)};
        \addplot[myorange] coordinates {(27.7,1)};
        \addplot[brightgreen] coordinates {(53.4,2)};
        \end{axis}
        
        \begin{axis}[
            width=.25\textwidth,
            xshift=3.0cm,
            yshift=-2.2cm,
            xbar=0pt,
            /pgf/bar shift=0pt,
            legend style={
            legend columns=4,
                at={(xticklabel cs:0.5)},
                anchor=north,
                draw=none
            },
            ytick={0,...,5},
            ytick style={draw=none},
            axis y line*=none,
            axis x line*=bottom,
            tick label style={font=\footnotesize},
            legend style={font=\footnotesize},
            label style={font=\footnotesize},
            xtick={0,20,...,100},
            bar width=3mm,
            yticklabels={},
            xticklabels={},
            xmin=0,
            xmax=100,
            area legend,
            y=4mm,
            enlarge y limits={abs=0.625},
            nodes near coords,
            nodes near coords style={text=black},
            every axis plot/.append style={fill}
        ]
        \addplot[lightpurple] coordinates {(21.8,0)};
        \addplot[myorange] coordinates {(78.2,1)};
        \addplot[brightgreen] coordinates {(0,2)};
        \end{axis}
        
        \begin{axis}[
            width=.25\textwidth,
            yshift=-6.0cm,
            xbar=0pt,
            /pgf/bar shift=0pt,
            legend style={
            legend columns=4,
                at={(xticklabel cs:0.5)},
                anchor=north,
                draw=none
            },
            ytick={0,...,6},
            ytick style={draw=none},
            axis y line*=none,
            axis x line*=bottom,
            tick label style={font=\footnotesize},
            legend style={font=\footnotesize},
            label style={font=\footnotesize},
            xtick={0,20,...,100},
            bar width=3mm,
            yticklabels={
                {USA},
                {UK},
                {Europe},
                {Canada},
                {Asia},
                {Oceania},
                {Unknown},
            },
            xticklabels={},
            xmin=0,
            xmax=100,
            area legend,
            y=4mm,
            enlarge y limits={abs=0.625},
            nodes near coords,
            nodes near coords style={text=black},
            every axis plot/.append style={fill}
        ]
        \addplot[lightpurple] coordinates {(9.9,0)};
        \addplot[myorange] coordinates {(1.2,1)};
        \addplot[brightgreen] coordinates {(1.8,2)};
        \addplot[myred] coordinates {(1.9,3)};
        \addplot[darkcyan] coordinates {(0.9,4)};
        \addplot[darkyellow] coordinates {(0.9,5)};
        \addplot[mypink] coordinates {(82.9,6)};
        \end{axis}
        
        \begin{axis}[
            width=.25\textwidth,
            xshift=3.0cm,
            yshift=-6.0cm,
            xbar=0pt,
            /pgf/bar shift=0pt,
            legend style={
            legend columns=4,
                at={(xticklabel cs:0.5)},
                anchor=north,
                draw=none
            },
            ytick={0,...,5},
            ytick style={draw=none},
            axis y line*=none,
            axis x line*=bottom,
            tick label style={font=\footnotesize},
            legend style={font=\footnotesize},
            label style={font=\footnotesize},
            xtick={0,20,...,100},
            bar width=3mm,
            yticklabels={},
            xticklabels={},
            xmin=0,
            xmax=100,
            area legend,
            y=4mm,
            enlarge y limits={abs=0.625},
            nodes near coords,
            nodes near coords style={text=black},
            every axis plot/.append style={fill}
        ]
        \addplot[lightpurple] coordinates {(63.8,0)};
        \addplot[myorange] coordinates {(5.9,1)};
        \addplot[brightgreen] coordinates {(8.8,2)};
        \addplot[myred] coordinates {(12.7,3)};
        \addplot[darkcyan] coordinates {(4.0,4)};
        \addplot[darkyellow] coordinates {(2.8,5)};
        \addplot[mypink] coordinates {(0.0,6)};
        \end{axis}
        
        \begin{axis}[
            width=.25\textwidth,
            yshift=-9.4cm,
            xbar=0pt,
            /pgf/bar shift=0pt,
            legend style={
            legend columns=4,
                at={(xticklabel cs:0.5)},
                anchor=north,
                draw=none
            },
            ytick={0,...,5},
            ytick style={draw=none},
            axis y line*=none,
            axis x line*=bottom,
            tick label style={font=\footnotesize},
            legend style={font=\footnotesize},
            label style={font=\footnotesize},
            xtick={0,20,...,100},
            bar width=3mm,
            yticklabels={
                {Atheist},
                {Christian},
                {Muslim},
                {Buddhist},
                {Hindu},
                {Unknown},
            },
            xmin=0,
            xmax=100,
            area legend,
            y=4mm,
            enlarge y limits={abs=0.625},
            nodes near coords,
            nodes near coords style={text=black},
            every axis plot/.append style={fill}
        ]
        \addplot[lightpurple] coordinates {(32.8,0)};
        \addplot[myorange] coordinates {(24.8,1)};
        \addplot[brightgreen] coordinates {(5.5,2)};
        \addplot[myred] coordinates {(2.2,3)};
        \addplot[darkcyan] coordinates {(0.8,4)};
        \addplot[darkyellow] coordinates {(33.8,5)};
        \end{axis}
        
        \begin{axis}[
            width=.25\textwidth,
            xshift=3.0cm,
            yshift=-9.4cm,
            xbar=0pt,
            /pgf/bar shift=0pt,
            legend style={
            legend columns=4,
                at={(xticklabel cs:0.5)},
                anchor=north,
                draw=none
            },
            ytick={0,...,5},
            ytick style={draw=none},
            axis y line*=none,
            axis x line*=bottom,
            tick label style={font=\footnotesize},
            legend style={font=\footnotesize},
            label style={font=\footnotesize},
            xtick={0,20,...,100},
            bar width=3mm,
            yticklabels={},
            xmin=0,
            xmax=100,
            area legend,
            y=4mm,
            enlarge y limits={abs=0.625},
            nodes near coords,
            nodes near coords style={text=black},
            every axis plot/.append style={fill}
        ]
        \addplot[lightpurple] coordinates {(50.3,0)};
        \addplot[myorange] coordinates {(35.6,1)};
        \addplot[brightgreen] coordinates {(9.3,2)};
        \addplot[myred] coordinates {(3.1,3)};
        \addplot[darkcyan] coordinates {(1.7,4)};
        \addplot[darkyellow] coordinates {(0.0,5)};
        \end{axis}
        
        \node[align=center,font=\small] at (2.7cm, 1.7cm) {\textbf{Gender}}; 
        \node[align=center,font=\small] at (1.2cm, 1.7cm) {2+Dem};
        \node[align=center,font=\small] at (4.1cm, 1.7cm) {4Dem};
        
        \node[align=center,font=\small] at (2.7cm, -0.6cm) {\textbf{Age}};
        \node[align=center,font=\small] at (1.2cm, -0.6cm) {2+Dem};
        \node[align=center,font=\small] at (4.1cm, -0.6cm) {4Dem};
        
        \node[align=center,font=\small] at (2.7cm, -2.8cm) {\textbf{Location}};
        \node[align=center,font=\small] at (1.2cm, -2.8cm) {2+Dem};
        \node[align=center,font=\small] at (4.1cm, -2.8cm) {4Dem};
        
        \node[align=center,font=\small] at (2.7cm, -6.6cm) {\textbf{Religion}};
        \node[align=center,font=\small] at (1.2cm, -6.6cm) {2+Dem};
        \node[align=center,font=\small] at (4.1cm, -6.6cm) {4Dem};
        
        \node[align=center,font=\small] at (2.7cm, -10.1) {Percentage of Users};
    \end{tikzpicture}
    }
    \caption{Distribution of the four demographic attributes in our two datasets, showing the set with all demographics known on the right and the random sample from those with at least two known on the left.}
    \label{fig:distribution_attributes}
\end{figure}

%% file: different_words_no_nn.tex
\begin{table*}
    \centering
    \small
    \bgroup
    \def\arraystretch{1.0}
    \setlength{\tabcolsep}{0.39em}
    \begin{tabular}{cccccccc}
        \toprule
        \multicolumn{2}{c}{Gender} & \multicolumn{2}{c}{Age} & \multicolumn{2}{c}{Religion} & \multicolumn{2}{c}{Location} \\
        Male & Female & Young & Old & Christian & Atheist & USA & Canada \\ \cmidrule(lr){1-2} \cmidrule(lr){3-4} \cmidrule(lr){5-6} \cmidrule(lr){7-8}
        \multicolumn{2}{c}{blush} & \multicolumn{2}{c}{health} & \multicolumn{2}{c}{embodying} & \multicolumn{2}{c}{america} \\ \cmidrule(lr){1-2} \cmidrule(lr){3-4} \cmidrule(lr){5-6} \cmidrule(lr){7-8}
        blushing & brow & regen & care & exalting & unionism & europe & original \\
        smile & eyeshadow & mana & reform & creaturely & mercantilist & country & tv \\
        chortle & bronzer & aid & healthcare & extols & american & canada & worst \\
        swoon & nars & permanent & education & mysteriousness & corporatocracy & sweden & hot \\
        snicker & nyx & condition & coverage & idolization & unfree & mexico & space \\
        wince & lipstick & treatment & high-deductible & magnanimity & proletarian & china & actual \\
        chuckle & mascara & mental & socialized & asceticism & environmentalist & india & body \\ 
        blushes & primer & preventative & medical & imbuing & wage-slavery & africa & home \\
        smirk & concealer & benefits & insurance & unalterable & communistic & usa & move \\
        guffaw & highlighter & medical & condition & mortification & free-marketeers & britain & nation \\
        \bottomrule
    \end{tabular}
    \egroup
    \caption{Examples of words with low overlap in nearest neighbors, showing how meaning can differ across the values of a demographic attribute.}
    \label{tab:word_differences}
\end{table*}

%% file: ppl_breakdown.tex
\begin{table}
    \centering
    \small
    \setlength{\tabcolsep}{5pt}
    \begin{tabular}{lr cc c cc}
        \toprule
                                                            &           & \multicolumn{2}{c}{2+Dem} & & \multicolumn{2}{c}{4Dem} \\
                                                            \cline{3-4} \cline{6-7}
        Att.                                                & Value     & 0D    & 4D    & & 0D    & 4D   \\
    \midrule
        \multirow{3}{*}{\rotatebox[origin=c]{90}{Age}}      & Young     & 107.1 & 103.6 & & 110.6 & 108.0 \\
                                                            & Old       & 115.1 & 111.1 & & 114.0 & 112.0 \\
                                                            & Unknown   & 112.3 & 108.6 & & -     & -     \\
    \midrule
        \multirow{7}{*}{\rotatebox[origin=c]{90}{Location}} & USA       & 108.5 & 105.7 & & 108.1 & 105.1 \\
                                                            & Canada    & 135.7 & 132.6 & & 110.0 & 107.6 \\
                                                            & Oceania   & 111.0 & 108.7 & & 114.8 & 112.8 \\
                                                            & Europe    & 130.0 & 128.2 & & 133.0 & 130.1 \\
                                                            & Asia      & 109.3 & 108.6 & & 145.3 & 145.4 \\
                                                            & UK        & 115.3 & 113.5 & &  96.9 &  96.9 \\
                                                            & Unknown   & 111.0 & 107.1 & & -     & -     \\
    \midrule
        \multirow{6}{*}{\rotatebox[origin=c]{90}{Religion}} & Christian & 116.5 & 111.9 & & 108.5 & 105.9 \\
                                                            & Atheist   & 106.4 & 103.2 & & 112.7 & 109.9 \\
                                                            & Muslim    & 112.7 & 108.5 & & 109.5 & 108.4 \\
                                                            & Hindu     & 122.4 & 115.6 & & 158.1 & 159.5 \\
                                                            & Buddhist  & 114.1 & 111.7 & & 116.4 & 114.1 \\
                                                            & Unknown   & 122.3 & 109.1 & & -     & -     \\
    \midrule
        \multirow{3}{*}{\rotatebox[origin=c]{90}{Gender}}   & Male      & 113.5 & 109.2 & & 115.2 & 112.6 \\
                                                            & Female    & 100.9 &  97.8 & & 102.7 & 100.4 \\
                                                            & Unknown   & 122.3 & 118.5 & & -     & -     \\
        \bottomrule
    \end{tabular}
    \caption{Perplexity for language models with no demographics (0D) or with all four demographic matrices (4D) with results broken down by demographic values.}
    \label{tab:ppl_demo_breakdown}
\end{table}

%% file: demo_aware_associations.tex
\begin{table*}[ht!]
    \centering
    \small
    \bgroup
    \def\arraystretch{1.0}
    \setlength{\tabcolsep}{0.39em}
    \begin{tabular}{rcccccccccccccccccc}
        \toprule
        & \multicolumn{2}{c}{best} & \multicolumn{2}{c}{oo3} & \multicolumn{2}{c}{oo10} & \multicolumn{2}{c}{best} & \multicolumn{2}{c}{oo3} & \multicolumn{2}{c}{oo10} & \multicolumn{2}{c}{best} & \multicolumn{2}{c}{oo3} & \multicolumn{2}{c}{oo10} \\
        Method & IN & US & IN & US & IN & US & M & F & M & F & M & F & Y & O & Y & O & Y & O \\
        \cmidrule(lr){1-1} \cmidrule(lr){2-7} \cmidrule(lr){8-13} \cmidrule(lr){14-19}
        C-SGM & 0.09 & 0.03 & 0.14 & 0.07 & 0.19 & 0.10 & 0.13 & 0.16 & 0.20 & 0.20 & 0.25 & 0.26 & - & - & - & - & - & - \\
        Ours G & \textbf{0.18} & \textbf{0.21} & \textbf{0.18} & \textbf{0.40} & 0.31 & 0.63 & 0.17 & 0.17 & \textbf{0.22} & 0.26 & 0.35 & 0.42 & 0.18 & 0.18 & \textbf{0.19} & 0.26 & 0.31 & 0.42 \\
        Ours G+D & 0.17 & 0.19 & 0.16 & 0.39 & \textbf{0.32} & \textbf{0.64} & \textbf{0.18} & 0.20 & \textbf{0.22} & \textbf{0.27} & 0.37 & \textbf{0.45} & \textbf{0.19} & \textbf{0.21} & \textbf{0.19} & \textbf{0.29} & \textbf{0.32} & \textbf{0.44} \\
        \bottomrule
    \end{tabular}
    \egroup
    \caption{Comparison of demographic-aware word association similarities for our embeddings using (G)eneric or (G)eneric+(D)emographic,  and the best results of the two variants of the composite skip-gram model (C-SGM) from \citet{garimella-etal-2017-demographic}. We show improved results for (US), (IN)dia, (M)ale, and (F)emale, and provide new results using age for (Y)ounger than 30 and (O)lder.}
    \label{tab:demo_associations}
\end{table*}

%% file: demo_aware_8subsets.tex
\begin{table*}
    \centering
\small
\begin{tabular}{llccccccccc}
\toprule
 & \hfill Gender & M & M & M & M & F & F & F & F & \\
 & \hfill Location & IN & IN & US & US & IN & IN & US & US & \\
Metric & Embeddings \hfill Age & Y & O & Y & O & Y & O & Y & O & Macro \\
\midrule
  \multirow{4}{*}{best} & Generic                 & \textbf{0.178} &        0.164   &        0.209   &        0.223   &        0.171   &        0.175   &        0.198   &        0.213   &        0.191   \\
                        & Age                     &        0.175   & \textbf{0.169} & \textbf{0.211} & \textbf{0.239} &        0.167   &        0.180   & \textbf{0.207} &        0.225   &        0.197   \\
                        & Age + Gender            &        0.175   & \textbf{0.169} & \textbf{0.211} & \textbf{0.239} & \textbf{0.174} & \textbf{0.181} & \textbf{0.207} & \textbf{0.227} & \textbf{0.198}$\dagger$ \\
                        & Age + Gender + Location \hspace{5mm} &        0.163   &        0.161   &        0.187   &        0.198   &        0.158   &        0.176   &        0.180   &        0.187   &        0.176   \\
         \midrule
  \multirow{4}{*}{oo3}  & Generic                 &        0.116   &        0.105   &        0.205   &        0.271   &        0.118   &        0.126   &        0.216   &        0.365   &        0.190   \\
                        & Age                     &        0.119   &        0.111   &        0.207   & \textbf{0.284} &        0.121   &        0.137   &        0.221   &        0.378   &        0.197   \\
                        & Age + Gender            &        0.120   &        0.111   &        0.207   & \textbf{0.284} &        0.123   &        0.136   & \textbf{0.235} &        0.378   &        0.199   \\
                        & Age + Gender + Location & \textbf{0.131} & \textbf{0.117} & \textbf{0.214} &        0.265   & \textbf{0.125} & \textbf{0.145} &        0.234   & \textbf{0.383} & \textbf{0.203}$\dagger$ \\
         \midrule
  \multirow{4}{*}{oo10} & Generic                 &        0.217   &        0.194   &        0.346   &        0.440   &        0.209   &        0.231   &        0.364   &        0.588   &        0.324   \\
                        & Age                     & \textbf{0.230} &        0.205   & \textbf{0.362} & \textbf{0.456} & \textbf{0.227} &        0.246   & \textbf{0.395} & \textbf{0.615} & \textbf{0.342} \\
                        & Age + Gender            & \textbf{0.230} &        0.205   & \textbf{0.362} & \textbf{0.456} & \textbf{0.227} & \textbf{0.250} &        0.389   &        0.614   & \textbf{0.342} \\
                        & Age + Gender + Location &        0.227   & \textbf{0.214} &        0.339   &        0.432   &        0.224   &        0.249   &        0.373   &        0.581   &        0.329 \\
\bottomrule
\end{tabular}
    \caption{Results on the 8 disjoint word association subsets for each combination of attributes. Similarities concatenate three embeddings that are each either generic, or specific to that demographic attribute. Overall, using age and gender in combination gives the best performance, though using all three is better on oo3. $\dagger$ indicates statistically significant improvement (permutation test, $p<0.001$) over the next best model on the marked metric.}
    \label{tab:demo_association_combinations}
\end{table*}

%% file: supplemental.tex
\section{Annotation of Demographic Attributes}
In order to verify the accuracy of our demographic attribute assignment, we manually annotated a sample of 100 users from the dataset.  Our extraction of attributes with regular expressions and rules was meant to have high-precision. It is likely that more attributes marked `unknown' by our extraction could be filled in upon manual inspection. We evaluate the retrieved attributes for these 100 users by viewing the set of all posts that matched our extraction rules and attempting to annotate age, religion, gender, and location. The annotation instructions were to identify the value of these four attributes based on the annotators interpretation of the text of the posts. Then, for cases where the extracted attribute is not `unknown', we calculate the percentage of times that they are the same. We get 94\% for location and gender, 78\% for religion, and 96\% for age. It should also be noted that despite the annotators best efforts, it is not possible to know the actual ground truth values.

\section{Reproducibility Criteria}
For each item in the list we have a section below with the relevant information.

\subsection{Experimental Results}
\paragraph{A clear description of the mathematical setting, algorithm, and/or model.}
The model we use is described in \citet{merity2018regularizing}. We modify it to support weight freezing and initialization. In Section 2 where they describe the weight-dropped LSTM, we concatenate our vectors for user-specific and demographic representations to $x_t$. 

The embeddings are obtained from the model described in \citet{bamman-etal-2014-distributed} for the demographic and user matrices. To obtain demographic vectors, we treat $C$, from Section 2, as a matrix whose rows represent the demographic attribute of a speaker (e.g. male, female) independent of the word used. The model updates the same way, changing a generic word vector and relevant demographic attribute vectors when backpropagating.

\paragraph{A link to a downloadable source code, with specification of all dependencies, including external libraries}
\begin{tightitemize}
    \item AWD-LSTM code is available from \url{https://github.com/salesforce/awd-lstm-lm}.
    \item Embedding code is available from \url{https://github.com/dbamman/geoSGLM}.
\end{tightitemize}

Code modifications will be available at \url{http://lit.eecs.umich.edu}. We use PyTorch 1.0.1 with CUDA 10.0.103.

\paragraph{Description of computing infrastructure used}
Each model is trained on one NVIDIA Tesla V100 GPU.

\paragraph{Average runtime for each approach}
Our methods take between 260 and 1450 seconds per epoch depending on the approach.

\paragraph{Number of parameters in each model}
The number of parameters for the model that uses all four demographic attributes has the most parameters at 249,752,492. Our smallest model is the user representation comparison which has 48,066,614.

\paragraph{Corresponding validation performance for each reported test result}
Validation perplexities are reported for the 2+Dem validation set. Table 5 validation perplexities:
\begin{tightitemize}
    \item demographic matrices ppl 62.57
    \item 500d baseline ppl 127.54
    \item 100d baseline ppl 124.39
    \item 100d generic ppl 111.70
    \item demographic age ppl 109.61
    \item demographic location ppl 109.93
    \item demographic gender ppl 109.68
    \item demographic religion ppl 110.88
\end{tightitemize}

Table 6 validation perplexities:
\begin{tightitemize}
    \item user vectors ppl 69.59
    \item user matrices ppl 62.11
    \item demographic matrices ppl 62.57
    \item generic ppl 65.44
\end{tightitemize}

\paragraph{Explanation of evaluation metrics used, with links to code}
Perplexity for language models is common and is implemented in Merity's code. The word association metrics for best, oo3, and oo10 are described in \citet{garimella-etal-2017-demographic} and we have reimplemented these metrics in order to compare to their results.

\subsection{Hyperparameter Search}
In our initial experiments on the 2+Dem validation set we chose the highest performing hyperparameters from the following list. One value listed means we used this value as described in Merity's code. Parameters were manually tuned and the best validation perplexity was chosen to use for all experiments.
\begin{tightitemize}
    \item embedding dimension: [100, 500] -- best 100
    \item LSTM hidden size (nhid): [550, 1150] -- best 1150
    \item wdrop: 0.0
    \item dropouti: 0.5
    \item dropouth: 0.5
    \item dropoute: 0.1
    \item embedding composition function: [addition, concatenation] -- best concatenation
    \item tied weights: [true, false] -- best false
    \item frozen pretrained embeddings: [true, false] -- best true
    \item LSTM layers (nlayers): 3
    \item learning rate: [10, 30] -- best 30
\end{tightitemize}

We also experimented with embedding dropout masks. We initially had separate masks for the generic and concatenated demographic-specific embeddings but if one is masked and not the other it doesn't mask all information about that word. When we tried embedding dropout with the same mask for each concatenated vector perplexity dropped several points.

The vocabulary size for demographic experiments was 502k, while the experiments for individual users had a vocabulary size of 177k words.

\subsection{Datasets}
\paragraph{Relevant statistics such as number of examples} See section 3 of the paper.
\paragraph{Details of train/validation/test splits} See Table 2 for details of the 2+Dem and 4Dem experiments and Section 5.2 for details on the user representation comparison.
\paragraph{Explanation of any data that were excluded, and all pre-processing steps} See Section 3.2.
\paragraph{A link to a downloadable version of the data} The Reddit data from 2007-2015 is available from \url{https://www.reddit.com/r/datasets/comments/3bxlg7/i\_have\_every\_publicly\_available\_reddit\_comment/}. Our subset of demographic labeled comments will not be available due to licensing restrictions, but can be reconstructed using our scripts and the source data linked to here. Our code can also be used to label more Reddit data from after this collection was posted.
\paragraph{For new data collected, a complete description of the data collection process, such as instructions to annotators and methods for quality control.} See Section 3 for data collection details and the beginning of this supplemental material for details on the annotation process.

